\documentclass[letterpaper, 10pt, conference]{ieeeconf}
\IEEEoverridecommandlockouts                              
\usepackage{times}
\usepackage{epsfig}
\usepackage{graphicx}
\usepackage{amsmath}
\usepackage{amssymb}
\usepackage{algorithmic}
\usepackage{booktabs} 
\usepackage[ruled]{algorithm2e}
\usepackage{diagbox}
\usepackage{xcolor}

\def\baselinestretch{0.955}

\usepackage[pagebackref=true,breaklinks=true,letterpaper=true,colorlinks,bookmarks=false]{hyperref}

\title{Physics-Based Dexterous Manipulations with Estimated Hand Poses and Residual Reinforcement Learning}

\author{Guillermo Garcia-Hernando$^{1,2}$, Edward Johns$^{1}$ and Tae-Kyun Kim$^{1,3}$
\thanks{$^{1}$Imperial College London, United Kingdom. $^{2}$Niantic, Inc., United Kingdom. $^{3}$KAIST, South Korea. This work was part of Imperial College London-Samsung Research project, supported by Samsung Electronics.}}

\begin{document}
\bstctlcite{IEEEexample:BSTcontrol} 

\maketitle

\begin{abstract}
Dexterous manipulation of objects in virtual environments with our bare hands, by using only a depth sensor and a state-of-the-art 3D hand pose estimator (HPE), is challenging. While virtual environments are ruled by physics, e.g. object weights and surface frictions, the absence of force feedback makes the task challenging, as even slight inaccuracies on finger tips or contact points from HPE may make the interactions fail. Prior arts simply generate contact forces in the direction of the fingers' closures, when finger joints penetrate virtual objects. Although useful for simple grasping scenarios, they cannot be applied to dexterous manipulations such as in-hand manipulation. Existing reinforcement learning (RL) and imitation learning (IL) approaches train agents that learn skills by using task-specific rewards, without considering any online user input. In this work, we propose to learn a model that maps noisy input hand poses to target virtual poses, which introduces the needed contacts to accomplish the tasks on a physics simulator. The agent is trained in a residual setting by using a model-free hybrid RL+IL approach. A 3D hand pose estimation reward is introduced leading to an improvement on HPE accuracy when the physics-guided corrected target poses are remapped to the input space. As the model corrects HPE errors by applying minor but crucial joint displacements for contacts, this helps to keep the generated motion visually close to the user input. Since HPE sequences performing successful virtual interactions do not exist, a data generation scheme to train and evaluate the system is proposed. We test our framework in two applications that use hand pose estimates for dexterous manipulations: hand-object interactions in VR and hand-object motion reconstruction in-the-wild. Experiments show that the proposed method outperforms various RL/IL baselines and the simple prior art of enforcing hand closure, both in task success and hand pose accuracy.
\end{abstract}
\vspace{-0.1cm}

\section{INTRODUCTION}
Capturing and transferring human hand motion to anthropomorphic hand models in physics-embedded environments, is the cornerstone of applications that require realistic interactions in VR/AR. 
To capture hand motion in such applications, most previous works resort to expensive and intrusive motion capture (mocap) systems, such as gloves \cite{rajeswaran2017learning}, exoskeletons and controllers \cite{zhang2017deep}. In this work, we aim to avoid such systems and explore a solution that allows us to perform dexterous manipulation actions by only using an estimate of the human hand pose.
\begin{figure}[!ht]
\begin{center}
 \includegraphics[width=1\linewidth,trim = {0.0cm 9cm 12.7cm 0.2 cm},clip]{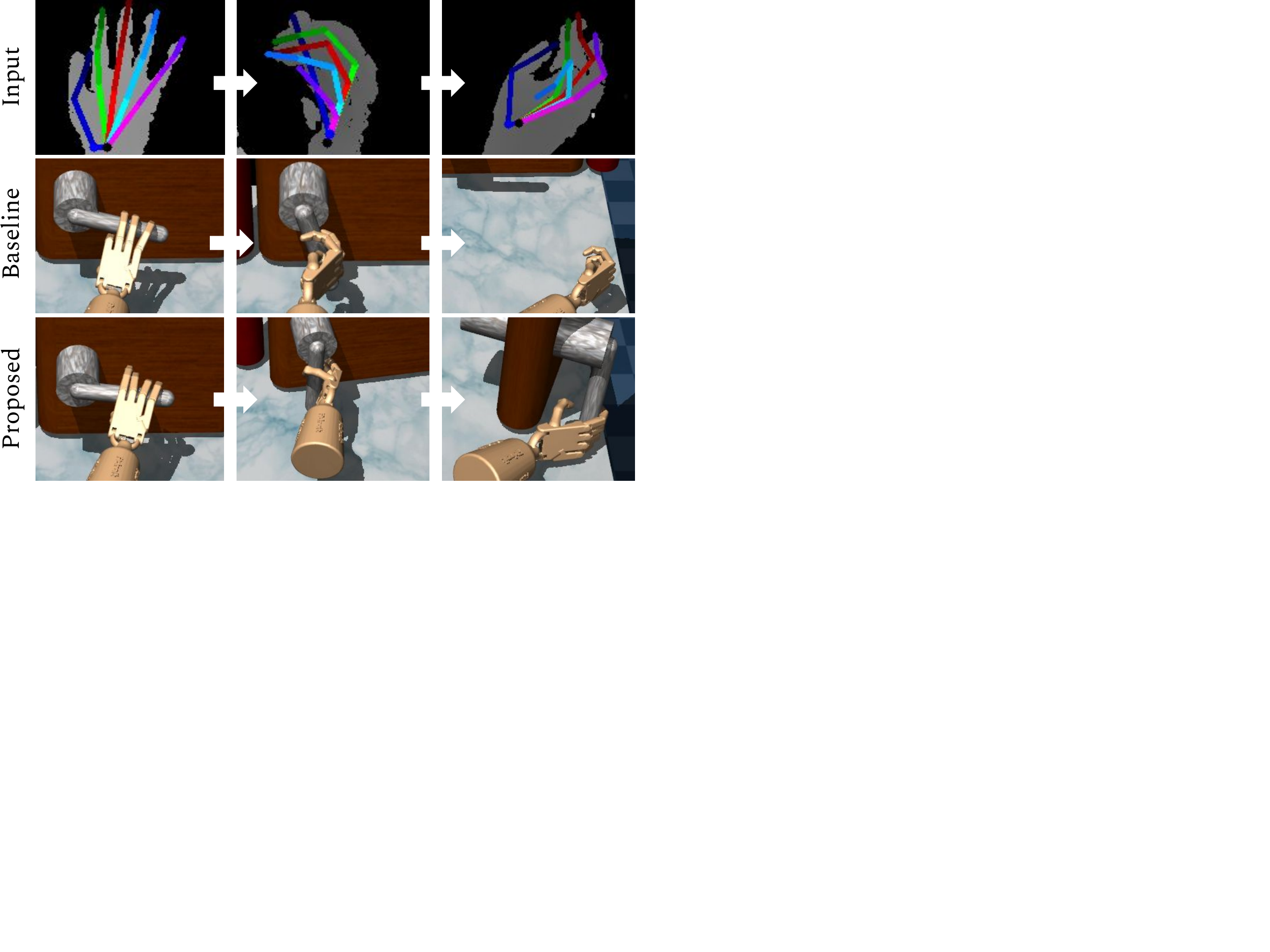}
\end{center}
\vspace{-0.45cm}
\caption{Mapping an estimated hand pose from a user, to a physically accurate virtual hand model is challenging. Simple pose retargeting functions fail due to the domain gap, contact physics, pose prediction errors and noise. Our method observes both the imperfect mapped hand pose from the user input, middle row, and the state of the simulation and produces a small residual correction that completes the task. To train our system, we generate input hand poses, top row, with a new data generation scheme that builds upon a mocap dataset~\cite{rajeswaran2017learning} and a large public hand pose dataset~\cite{yuan2017bighand2}. Note that the depth camera is pointing to the human hand from the ground.} 
\vspace{-0.7cm}
\label{fig:teaser}
\end{figure}

Hand pose estimators (HPEs) typically produce 3D locations of keypoints of a human hand model. 
Given the difference between the human hand and the hand model, the design of a function mapping an input hand pose to the model's parameters is needed, a process known as inverse kinematics or motion/pose retargeting.
Designing a function that produces a visually similar output is relatively straightforward, and hand-engineered \cite{leapmotion, li2018vision}, data-driven \cite{oikonomidis2011efficient, khamis2015learning} or hybrid \cite{li2018vision, antotsiou2018task} solutions are available. 
However, when interacting with the simulated physical environment, 
visual resemblance between input and target is not enough, given that one needs to consider both contact physics between the hand and object and input noise coming from the hand pose estimator as shown in Fig.~\ref{fig:teaser}. 
Commercial solutions~\cite{leapmotion, hololens} circumvent these problems simply by ignoring physics laws and `attracting' the hand towards the object. 

Other approaches model the underlying contact physics by establishing relationships between the virtual penetration of the hand on the object \cite{holl2018efficient}.
Such solutions, despite being effective for some simple grasping actions, do not produce physically realistic motion in the target domain. Also, the inferred contact force will depend directly on the noisy pose estimate, making it difficult to apply the precise forces and subtle movements required in some dexterous tasks. 

Related to our work, \cite{Tzionas:IJCV:2016,hasson19_obman} track and reconstruct 3D hand-object interactions using simple physics constraints such as contact and mesh penetration. In contrast, we generate complete physics-aware sequences using a physics simulator, which can actually succeed in the task of interest. Related to us, and aiming to generate physically plausible sequences from vision, \cite{peng2018sfv, yuan2019ego} use RL for full body poses. Different to~\cite{peng2018sfv}, which aims to teach an agent to autonomously perform by observing a single reconstructed and filtered video, our work aims to correct noisy user hand poses `as they come', and \textit{assist} the user in a similar setting to shared autonomy~\cite{reddy2018shared}. In contrast to \cite{yuan2019ego}, which aims to estimate the ego-pose of the humanoid by indirectly observing from their character point of view, we directly observe user's hand motion and assist in achieving the task while generating virtual poses similar to the visual input.

\begin{figure*}[!htp]
\vspace*{0.2cm}
\begin{center}
\  \includegraphics[width=1\textwidth, trim = {6.6cm 8.1cm 11.5cm 6.1cm},clip]{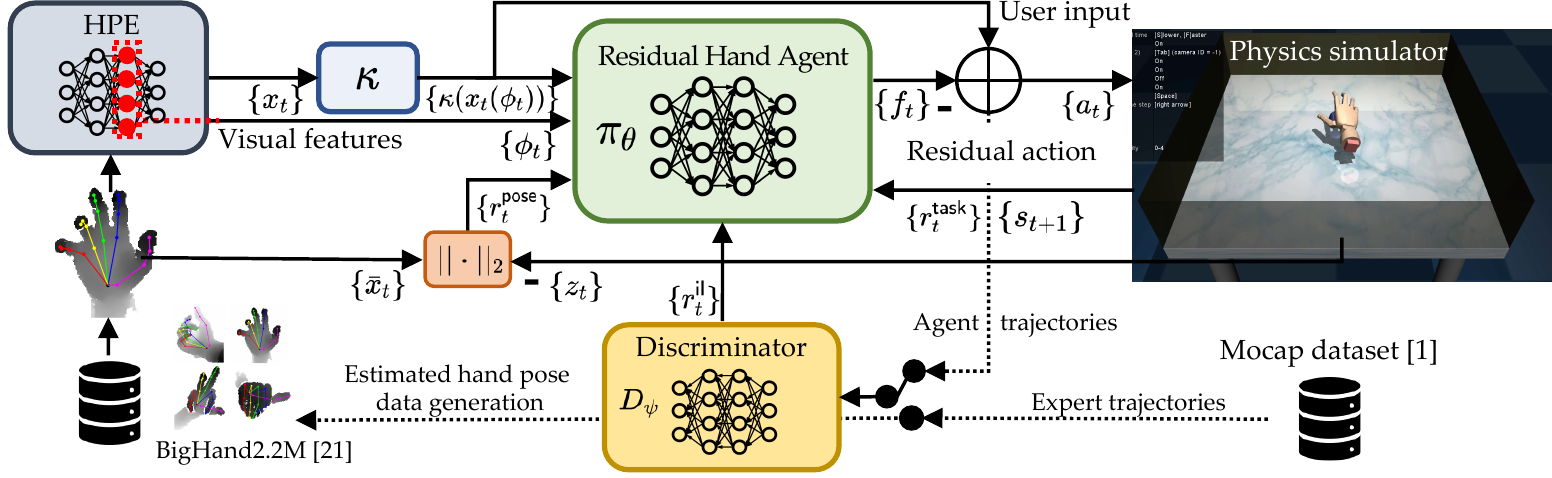}
\end{center}
\vspace{-0.55cm}
\caption{\textbf{Framework training overview.} During training, the residual agent performs actions that aim to correct the user input, and receives feedback from both the simulator and a discriminator. The discriminator indicates how much the actions resemble expert human actions from a mocap dataset~\cite{rajeswaran2017learning}, whilst the simulator allows us to generate several samples of rich physics simulation and to measure the resemblance between input and virtual poses. To train our framework in the absence of ground-truth pairs hand poses and actions, we can generate estimated hand poses by finding images on a large hand pose dataset~\cite{yuan2017bighand2} that are likely to have generated the actions from the mocap dataset. Once we find these samples, we pass them through a hand pose estimator and an inverse kinematics or pose retargeting function to generate user input. Algorithm details can be found in the Appendix.}
\label{fig:training}
\vspace{-0.4cm}
\end{figure*}

We propose a system, illustrated in Fig.~\ref{fig:training}, that observes an imperfect user input and refines it in order to accomplish the manipulation task. We define the user input, Section \ref{subsec:IK}, as an estimated hand pose mapped by an inverse kinematics or pose retargeting function. To achieve this, we introduce a residual agent that acts on top of the user input in Section \ref{subsec:RHA}. We assume that the user input is similar to the optimal action --modern HPEs present average joint errors in the range of 7 to 15 mm \cite{yuan2018depth}-- and only require a correcting stage to produce the correct kinematics. In order to automatically learn this correction without making any assumptions on the underlying contact physics, we train the residual agent using reinforcement learning (RL) in a model-free setting \cite{schulman2017proximal, openai} within an accurate physics simulator \cite{todorov2012mujoco}. To avoid unnatural motion typically present under RL framework \cite{rajeswaran2017learning}, our system builds upon recent work in adversarial imitation learning (IL) \cite{ho2016generative,zhu2018reinforcement}, that uses a discriminator to encourage the policy to produce actions similar to trajectories from a dataset captured using a mocap glove~\cite{rajeswaran2017learning}. 
Unlike prior arts~\cite{leapmotion, hololens,holl2018efficient}, our method enables dexterous manipulations e.g. in-hand pen manipulation or picking a coin. The proposed residual agent is also learned by the 3D hand pose estimation reward, improving HPE accuracy when the physics-guided corrected target poses are re-mapped to input space. These objectives are presented in Section \ref{subsec:rewards}. 

To train such a framework, we need continuous intended action sequences of noisy estimated hand poses, as well as some successful manipulation actions obtained by mocap data. It is difficult to collect such HPE sequences in an online fashion, because users tend to stop their motions in the middle of the tasks when they fail.
We first explore generating noisy input sequences by adding random noise to the ground-truth mocap data. To circumvent the gap between the synthetic noise and the real structured noise coming from HPE, we propose, in Section \ref{subsection:data_gen}, a data generation approach which, given a dataset of successful manipulation sequences in the virtual space  \cite{rajeswaran2017learning}, finds a ground-truth hand pose and depth image that is most likely to have generated such action, by querying a public large scale hand pose dataset \cite{yuan2017bighand2}. Using this pipeline we conduct experiments on two potential applications of our framework. The first one, \textit{Experiment A}, appears in Section \ref{exp:a} and it studies a typical VR scenario where the user interacts with the environment with their bare hands in mid-air and a hand pose estimator. In the second one, \textit{Experiment B} in Section \ref{exp:B}, we aim to reconstruct in a physics simulator hand-object RGBD sequences captured in-the-wild with the use estimated hand poses and initial object pose estimates.
In various experiments, our proposed method outperforms RL/IL baselines, and some relevant arts. 
\section{RELATED WORK}
\vspace{-0.2cm}
\textbf{3D hand pose estimation }consists of estimating the 3D locations of hand keypoints given an image. A main part of the success in the field comes from the use of depth sensors \cite{oikonomidis2011efficient, keskin2012hand, tang2014latent} and deep learning \cite{tompson2014real, oberweger2015hands,  yuan2018depth}, while recent successful approaches exploit single RGB images as input~\cite{zimmermann2017learning, GANeratedHands_CVPR2018}. Note that most current hand pose estimators only output 3D joint locations than angles, making the mapping between locations and angles not trivial; however there is some promising work on estimating 3D hand meshes that could make this problem easier \cite{romero2017embodied, boukhayma20193d}. 
 \\                                     
\indent\textbf{Vision-based teleoperation.} Traditionally, teleoperation has been limited to mapping the human hand to the (physical or virtual) robot hand by using contact devices such as tracking sensors \cite{cerulo2017teleoperation}, exoskeletons \cite{borst2005realistic} and gloves \cite{kumar2015mujoco}. Some vision-based approaches exist \cite{kofman2007robot,romero2011human, du2012markerless, li2018vision, antotsiou2018task, handa2019dexpilot} but are limited to simple grasping actions. \cite{li2018vision} proposes a retargeting method between depth images and a robotic hand model, however the mapping function is purely based on hand appearance ignoring objects. \cite{antotsiou2018task} combines inverse kinematics with a PSO function that encourages contact between object and hand surfaces. We share with \cite{antotsiou2018task} the aim of achieving realistic interactions, but simply forcing contact is not enough for dexterous actions such as in-hand manipulation. \cite{handa2019dexpilot} introduces a HPE tailored to a robot hand model. Given that our framework is HPE-agnostic, both works are complementary and could to produce a solid system if combined. In the VR and graphics community, perhaps the simplest approach for tackling such problems, and as adopted by commercial products such as Leap Motion~\cite{leapmotion} or Hololens~\cite{hololens}, is to recognize the ongoing hand gesture, e.g. swipe or pinch, and then trigger a prerecorded output \cite{ buchmann2004fingartips, moehring2011effective, yim2016gesture}. However, such approaches produce artificial motion that often deviates significantly from the user input. Similarly, the interaction engine by Leap Motion~\cite{leapmotion} recognizes the gesture and `attracts' the object to the hand producing an artificial `sticking' effect. Our method corrects the user input slightly, but only enough to achieve the task, and importantly it respects the laws of physics. Other works use a priori information about the hand and the scene, by synthesizing a grasp from a predefined database \cite{rijpkema1991computer, elkoura2003handrix, miller2003automatic, li2007data, prachyabrued2012virtual},~limited to a specific set of objects and interactions, and very sensitive to uncertainty about the environment. Some works attempt to model the contact physics~\cite{kry2006interaction, liu2009dextrous, ye2012, zhao2013robust, kim2015physics, holl2018efficient} to infer contact forces between the hand and objects, by measuring, for example, the penetration of the user hand into the object mesh. The main problem of such approaches is that the computed contact force relies on high-precision hand pose estimation, and the method tends to apply forces that do not necessarily transfer to the real world without unexpected consequences. 

\textbf{Physics-based pose estimation.} \cite{Tzionas:IJCV:2016} uses a physics simulator within an optimization framework to refine hand poses, following earlier generative and discriminative model fitting work~\cite{hamer2009tracking, kyriazis2013physically, kyriazis2014scalable,rogez2015understanding}. \cite{hasson19_obman} presents an end-to-end deep learning model that exploits a contact loss and mesh penetration penalty similar to \cite{Oikonomidis2011, kyriazis2013physically, Tzionas:IJCV:2016,Tsoli2018}, for plausible hand-object mesh reconstruction. These estimators are subject to simple physical constraints such as contact and mesh penetration and deal with single-shot images. In~\cite{yuan2019ego}, physically-valid body poses are estimated and forecasted from egocentric videos using RL. Their aim is to estimate the ego-pose of the humanoid by indirectly observing from their character point of view using similar rewards as \cite{peng2018sfv}, discussed below. 

\textbf{Motion retargeting and reinforcement learning.} 
Our problem shares similarities with full body motion retargeting \cite{geijtenbeek2013flexible}, particularly with methods that consider accurate physics on the target space and train control policies using RL~\cite{heess2017emergence,peng2018deepmimic,chentanez2018physics,liu2018learning, peng2018sfv}. \cite{peng2018deepmimic, chentanez2018physics, liu2018learning} propose an RL approach to learn skills from a reference mocap motion. \cite{peng2018sfv} extends such work to deal with reference motion from a body pose estimation step that is cleaned and post-processed to mimic the motions, as in \cite{peng2018deepmimic}. The main difference of our work is that we perform online predictions given a noisy user input instead of learning to mimic a skill in an offline fashion. 
For this reason, we embrace the noisy nature of our problem and propose the residual learning guided by the hand pose estimation reward and the noisy data generation scheme. 

\textbf{Robot dexterous manipulation and reinforcement learning.} 
For attempting to learn robotic manipulation skills without user input, and using both RL and IL, we highlight three recent works~\cite{zhu2018reinforcement, rajeswaran2017learning, openai}. We share with \cite{zhu2018reinforcement} a similar adversarial hybrid loss, however our model has significantly more degrees of freedom.
We build upon \cite{rajeswaran2017learning}'s simulation framework, using their dataset of glove demonstrations, and extend the environments to deal with vision-based hand pose estimation. We share with \cite{openai} the ambition of learning physically accurate dexterous hand manipulations, but more in physics embedded VR space using user's hand via state-of-the-art hand pose estimator.

\textbf{Residual policy learning.} We discuss two recent papers proposing a similar residual policy idea~\cite{johannink2018residual, residual-mit}. We share with these works the residual nature of our policy and the idea that improving an action, instead of learning from scratch, significantly helps the exploration problem of RL and tends to produce more robust policies. The main difference from our work is that our residual action works on top of a user input instead of a pre-trained policy, i.e. our policy observes the action taken by the user and the world and then acts accordingly, instead of just observing the state of the world, which could lead to a discrepancy between the user's intention and the agent. Other differences include the nature of the problem, the complexity of the action space, the combination with adversarial IL, and a problem setting similar to shared autonomy \cite{reddy2018shared}.
 \vspace{-0.1cm}
\section{PROPOSED FRAMEWORK}
\subsection{Inverse kinematics: from human hand pose to virtual pose}\label{subsec:IK}
Given the user's estimated hand pose $x_t$, which consists of the 3D locations of 21 joints of a human hand \cite{yuan2017bighand2} on a given visual representation $\phi_t$ at time step $t$, we aim to obtain a visually similar hand posture $z_t$ in our virtual model. This requires estimating parameters $a_t$, defined as the actuators or \textit{actions} of the virtual hand model which determine the target angle between hand joints with the help of PID controllers. 

Inverse kinematics (IK) refers to the task of computing rotations $a_t$ such that the virtual hand pose $z_t$ is equivalent to the user's hand pose $x_t$. Note that $z_t$ belongs to a different domain to $x_t$, but it can be measured by carefully placing sensors in the virtual hand model. This mapping from pose to rotations, $\kappa$, can be manually designed or automatically learned, for example with a supervised neural network when input-output pairs are available, and can be written as: 
\vspace{-0.2cm}
\begin{equation}
\label{eq:ik}
    a_t = \kappa(x_t(\phi_t)).
\end{equation}
For simplicity, we often refer to $\kappa(x_t(\phi_t))$, in the action space, as the \textit{user input}, in contrast to user's estimated hand pose $x_t$, in the pose space. IK is inherently an ill-posed problem, since depending on how different the virtual and human models are, the target pose $z_t$ can potentially be reached by multiple $a_t$'s, or there may not be a solution at all. This problem becomes even more aggravated when the input $x_t$ is noisy, which is the nature of a hand pose estimator. We describe our residual approach to deal with this imperfect input next.
\subsection{Residual Hand Agent} \label{subsec:RHA}
We now describe how to train the residual controller, which acts upon the output of the above IK function. Due to both the imperfect mapping between the human kinematics and virtual kinematics, and the noise introduced by the hand pose estimator, we assume that the user input $\kappa(x_t(\phi_t))$ produces actions that are \textit{close} to optimal, but not sufficiently good to succeed in the task of interest. As an additional requirement for optimal action predictions, the temporal nature of our sequences means that a small early mistake can later have a catastrophic effect due to compounding errors that propagate to subsequent simulation stages. The residual controller introduces a \textit{residual} action $f_t$, which is a function of $\kappa(x_t(\phi_t))$, the current simulation state $s_t$ and the visual representation $\phi_t$, which can be either an image or extracted visual features. Those terms are combined as follows:
\begin{equation}
\label{eq: action_res}
    a_t = \kappa(x_t(\phi_t))-f_t(s_t, \kappa(x_t(\phi_t)), \phi_t). 
\end{equation}
In order to not deviate from the user input significantly, we limit the residual action $f$ to be within a certain zero-centered interval. We formulate the learning of the residual policy as a RL problem, where an agent interacts with a simulated environment by following a policy $\pi_\theta(f|s, \kappa, \phi)$ parametrized by $\theta$, which in our case is a neural network. 
 
 The state $s$ includes the current information tailored to every task of the simulation environment, such as the relative positions between the target object and the virtual hand model, the model's velocity, etc. At each time step $t$ the agent observes $s_t$, $\kappa(x_t(\phi_t)$ and $\phi_t$, samples an action $f_t$ from $\pi_\theta$, and an action $a_t$ is applied to the environment.  The environment moves to the next state $s_{t+1}$ sampled from the environment dynamics, which we assume to be unknown. A scalar reward $r_t$ quantifies how good or bad this transition was, and thus our goal is to find an optimal policy that maximizes the expected return, defined as $
J(\theta) = \mathbb{E}_{\tau \sim p_\theta(\theta)} \left[\sum_{t = 0}^T \gamma^t r_t \right]$, where $p_\theta(\tau)$ 
is the distribution over all possible trajectories $\tau = (s_0, \kappa(x_0), \phi_0, f_0, s_1, ...)$ following the policy $\pi_\theta$. The term $\sum_{t = 0}^T \gamma^t r_t$ represents the total return of a trajectory for a horizon of $T$ time steps and a discount factor $\gamma \in [0,1]$. In our problem, $T$ is variable depending on the length of the hand pose input sequence. State and rewards details can be found in the Appendix.

To optimize $\theta$ several methods can be used, however in this work we focus on a popular policy gradient approach proximal policy optimization (PPO) \cite{schulman2017proximal} due to its recent success on learning dexterous policies without user input \cite{openai}. This approach optimizes $J$ over $\theta$ to maximize the return. The gradient of the expected return $\nabla_\theta J(\theta)$ is estimated with trajectories sampled by following the policy, and learns both a policy network and a value function, which estimates the expected return when following the policy.

\subsubsection{Reward function}\label{subsec:rewards}
The total reward function $r_t$ that guides the framework learning process is defined as:
\begin{equation}
 r_t=  \omega^{\mathsf{task}}r^{\mathsf{task}}_t+ \omega^{\mathsf{il}}r^{\mathsf{il}}_t+\omega^{\mathsf{pose}}r^{\mathsf{pose}}_t,
\end{equation}
where $\omega^{\mathsf{task}}$, $\omega^{\mathsf{il}}$ and $\omega^{\mathsf{pose}}$ are weighting factors.

\paragraph{Task-oriented reward} $r_t^{\mathsf{task}}$: it is tailored for each environment and guides the policy towards desirable behaviours in terms of task accomplishment, with short-term rewards such as getting close to the object of interest, and long-term rewards such as opening the door (see Appendix). 

\paragraph{Imitation learning reward} $r^{\mathsf{il}}_t$: Policies learned with only RL tend to produce unnatural behavior: they are effective to accomplish the task of interest, but produce actions that a human would never do~\cite{rajeswaran2017learning}. To encourage action sequences that more closely resemble expert data, we add the following adversarial IL reward function similar to~\cite{zhu2018reinforcement}: \setlength{\abovedisplayskip}{3pt}
\begin{equation}
    r^{\mathsf{il}}_t = (1-\lambda )\log(1-D_\psi(s_t,a_t)),
\end{equation}
where $D_\psi$ is a score quantifying how good an action is, given by a discriminator with parameters $\psi$. To include this objective in our framework, we use a min-max objective \cite{ho2016generative}:
\begin{equation}
\min_{\theta} \max_\psi \mathbb{E}_{\pi_{E}}[\log D_\psi(s,a)] + \mathbb{E}_{\pi_\theta}[\log(1-D_\psi(s,a))],
\end{equation}
where $\pi_{E}$ denotes an expert policy generated from demonstration trajectories. This objective encourages the policy $\pi_\theta$ to produce actions $f_t$ that correct the user input $\kappa(x_t(\phi_t))$, generating pairs of $(s_t, a_t)$ that are similar to those of an expert. In our framework, we obtain $\mathcal{D} = {(s_i,a_i)}_{i=1...N}$ from \cite{rajeswaran2017learning}, which used a data glove and a tracking system \cite{kumar2015mujoco} to capture noise-free sequences. 
\paragraph{3D hand pose estimation reward} $r^{\mathsf{pose}}$:
The reward terms introduced above can lead to virtual poses $z_t$, that diverge from the pose depicted on the user input image, particularly if the hand pose estimator fails due to object occlusion. If we have access to annotated ground-truth hand poses $\bar{x}_t$ during training, we can introduce an additional reward that encourages the policy network to produce actions that visually resemble the user pose and is defined as:
\vspace{-0.1cm}
\begin{equation}
r^{\mathsf{pose}}_t =  -\sum_j^{21}||z^j_t-\bar{x}^j_t||_2,
\end{equation}\vspace{-0.1cm}
where $z^j_t$ and $\bar{x}^j_t$ denote the 3D position of the $j$-th joint of the human and model respectively.
\vspace{-0.1cm}
\subsection{Data generation scheme}
\label{subsection:data_gen}
If we examine Eq. \ref{eq: action_res} we observe that, to train our residual policy, we need a dataset of estimated hand poses $\{x_t\}$ depicting natural hand motion that would produce a successful interaction if the system was perfect. We could think of recording hand pose sequences by asking users to perform the action `as if it was successful', but given the temporal dependency of the problem we would be acquiring data somewhat different from the true distribution. 

Our idea consists of using a mocap dataset which contains successful sequences of state-action pairs and find hand images that could have produced these actions by querying a 3D hand pose dataset. For this approach to work, a dense and exhaustive 3D hand pose dataset in terms of articulations and relative camera-hand viewpoints is needed. We use BigHand2.2M~\cite{yuan2017bighand2} as hand pose dataset and the dataset introduced in Rajeswaran et al. \cite{rajeswaran2017learning} as mocap dataset. We first measure the virtual poses $\{z_t\}$ generated by the actions by placing virtual sensors and a virtual camera. Given the sequences of virtual poses, we retrieve the closest ground-truth poses in a 3D hand pose dataset. We tried different representations and query functions for retrieval, but got the best results by retrieving similar viewpoints and later refining by the distance of aligned and palm-normalised joint coordinates. Once the matches are found, we retrieve their associated image and compute estimated hand poses by passing the images through a 3D hand pose estimator.

\begin{figure*}[!htb]
\begin{center}
\vspace*{0.2cm}
  \includegraphics[width=1\textwidth, trim = {0cm 13.4cm 14.3cm 0.1cm},clip]{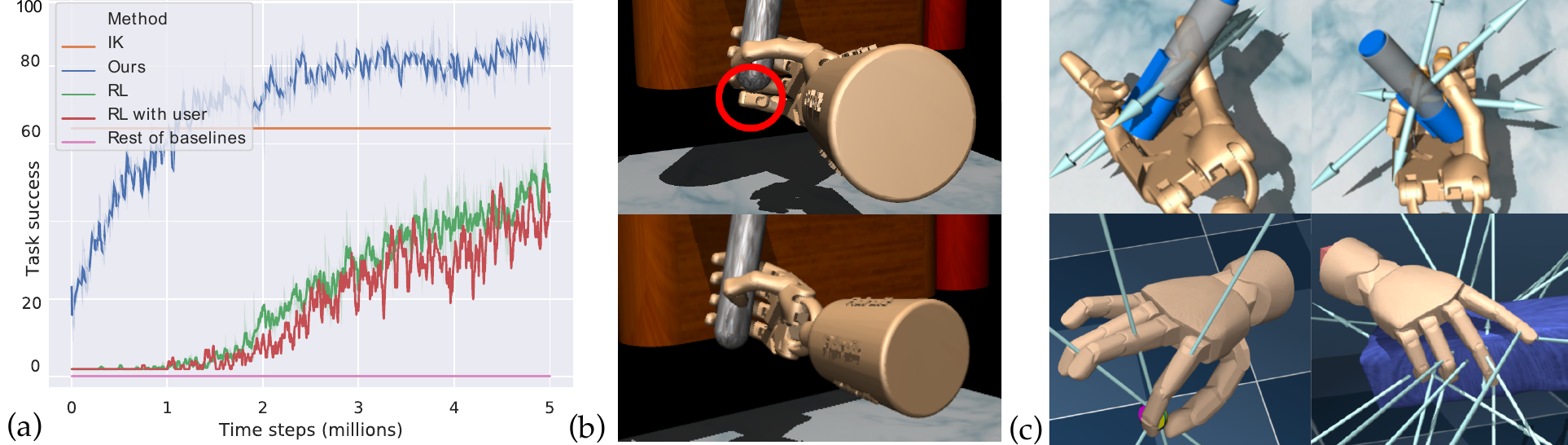}
\end{center}
\vspace{-0.55cm}
\caption{Training curves on `opening door' for our approach and baselines (a). (b) Qualitative ablation study on reward function (top) Our agent with only task reward $r^{\mathsf{task}}_t$ and (bottom) adding $r^{\mathsf{il}}_t$ on the same input sequence with equal weights. (c) Resulting contact forces for in-hand manipulation, `give coin' and `pour juice'. For in-hand manipulation, approaches maximizing  contact cannot accomplish the task. }
\label{fig:exp1}
\vspace{-0.6cm}
\end{figure*}
\vspace{-0.15cm}
\section{EXPERIMENTS\protect\footnote{Appendix can be found at the end of this document and videos in the project page: \url{https://sites.google.com/view/dexres}}}
\vspace{-0.1cm}
\subsection{Performing dexterous manipulations in a virtual space with estimated hand poses in mid-air}\label{exp:a}
\vspace{-0.1cm}
In this experiment we evaluate our framework when we have access to a glove-recorded mocap dataset~\cite{rajeswaran2017learning} with successful expert trajectories and we use our data generation scheme. As HPE we use \cite{ye2016spatial} and to train and retrieve images with particular poses we use BigHand2.2M dataset~\cite{yuan2017bighand2}, which was designed to densely capture articulation and viewpoint spaces in mid-air and in an object-free setup. Because of the absence of object occlusions in BigHand2.2M, we drop $r^\mathsf{pose}$ and do not feed visual features to the policy network.  We first evaluate our framework in a controlled setting where we add synthetic noise to expert demonstrations and then we evaluate it with real structured hand pose estimation noise.  \\
\textbf{Hand model:} We use the ADROIT anthropomorphic platform \cite{rajeswaran2017learning}, consisting of 24 degrees-of-freedom (DoF) joint angle rotations of Shadow dexterous hand, plus 6 DoF defining the 3D position and orientation of the hand.\\
\textbf{Simulator and tasks:} We use the MuJoCo physics simulator \cite{todorov2012mujoco} and the four dexterous manipulation scenarios defined in \cite{rajeswaran2017learning}: door opening, in-hand manipulation, tool use and object relocation. In `door opening' the task is to undo the latch and swing the door open. In `in-hand manipulation' the task is reposition a blue pen to match the orientation of a target pose (green pen). `Tool use': the task consists of picking up a hammer and drive the nail into a board. `Object relocation' aims to move a blue ball to a green target location. Each task is considered successful if the target is achieved with a certain tolerance. There are about 24 mocap trajectories per task and we split them in equal training-test sets. \\
\textbf{Policy network:} $\pi$ is a (64, 64) MLP and the residual policy is limited to 20\% of the action space. The action is modeled as Gaussian distribution with a state-dependent mean and a fixed diagonal covariance matrix. We use the same architecture for value function and discriminator.\\ 
\textbf{Baselines.} In this experiment we evaluate the following:\\
\noindent\textbf{Inverse kinematics (IK):} The action applied is based solely on user's input and we specify below its nature. \\
\textbf{Reinforcement learning (RL):} The agent observes both the user input and the state in a non-residual way \cite{schulman2017proximal} without access to demonstrations. Two versions: `RL - no user' with only task reward and `RL + user reward' with additional reward term encouraging following the user input.\\
\textbf{Imitation learning (IL):} The agent observes both the user input and the state and it has access to demonstrations during the adversarial learning process based on GAIL \cite{ho2016generative}.\\
\textbf{Hybrid learning:} We combine the above baselines in a similar way to our proposed algorithm without residual.\\
Implementation details, states and rewards definitions, and training parameters can be found in the Appendix. 
\subsubsection{Overcoming random noise on demonstrations}
The aim of this experiment is to verify whether our framework can deal with noisy observations and produce useful residual actions. In this scenario we have total control on the amount and nature of the noise allowing us to dissect the results. In this experiment the user inputs are the expert successful actions recorded using a mocap glove from \cite{rajeswaran2017learning} on the `opening door'
 environment, thus we can assume they are free of noise. We synthesize noise by adding a zero-mean Gaussian noise with standard deviation $\sigma$ radians to each actuator, on top of the user input in both training and test trajectories. Note that errors in a single actuator propagate through the linked joints by forward kinematics.

After training a policy for a certain $\sigma$, we show its generalization to other values of noise on test sequences in Table \ref{table:sigmas_exp1}. We observe that our residual agent is able to recover meaningful motion up to a $\sigma_{\mathsf{test}}$ of 0.20 rad when similar magnitudes have also been observed in training. The noisy user input can succeed a significant amount of times alone provided that small changes, or changes in the right direction, may not affect the overall success.
\vspace{-0.1cm}
\begin{table}[h]
\caption{Our approach for different noise levels in training/test}
\vspace{-0.4cm}

\label{table:sigmas_exp1}
\centering
\resizebox{0.95\columnwidth}{!}{%
\begin{tabular}{ccccccc}
& \multicolumn{6}{c}{$\sigma_{\mathsf{test}}$} \\
\cmidrule(lr){2-7}
$\sigma_{\mathsf{train}}$   &
$ 0.00$    &
$ 0.01$   &
$ 0.05$   &
$ 0.10$   &
$ 0.15$   &
$ 0.20$   \\
 \midrule
0.01 & 71.00  &  70.00 & 52.00  & 26.00 & 9.00  & 1.00 \\
0.05 & 100.0  &  90.00 & 83.00 & 50.00 & 24.00 & 4.00 \\
0.10 &  91.00 &  89.00 &  87.00 & \textbf{87.00} & 56.00 & \textbf{26.00} \\
0.15 & \textbf{100.0} & \textbf{ 96.00 } & \textbf{92.00} & 80.00 & \textbf{57.00} & 19.00 \\
0.20 & 71.00 &  74.00 & 75.00  & 71.00  & 47.00  & 20.00 \\\midrule
User input: & 80.00 & 86.30 & 74.00  & 33.80 & 9.20 & 2.70 \\
\bottomrule
\end{tabular}
 }
\end{table}
\vspace{-0.1cm}

In Table \ref{table:baselines_exp1} we show the performance of different baselines for a fixed $\sigma$ value of $0.05$ rad for each environment. Two results are reported: the task success on noisy sequences generated from demonstration sequences that are only seen during training, and the accuracy on an independent test set. Some of the baselines do not succeed on some environments, being consistent with the results reported by \cite{rajeswaran2017learning}. Furthermore, when training our residual policy, it converges significantly faster than other baselines (see Fig. \ref{fig:exp1} (a)). For instance, our policy converges after 3.8M and 5.2M samples for door opening and in-hand manipulation, compared to 7.9M and 13.8M for RL baseline. In our RL framework for our approach and baselines, 5M samples with network updates are generated in about 12 hours on a single core machine with a GTX 1080Ti, while RL alone baselines require only. The reason for this faster convergence is the help in exploration that the user input brings to the learning process~\cite{residual-mit}.  For the last scenario, `object relocation', none of our baselines nor our approach is able to correct the user input and degrade its performance. We hypothesize that the low result of PPO propagates to our algorithm and using other optimization could help to recover the user input \cite{rajeswaran2017learning}. 

To conclude this experiment we perform an ablation study to evaluate the impact of each RL and IL components of our approach. Combining both leads to accomplishing the task while keeping a motion that resembles the human experts more closely (see Fig. \ref{fig:exp1} (b) and video). In terms of task success, RL alone achieves 75.9\% while IL alone 36.5\%. 
\begin{table}[t]  
\vspace{0.2cm}
\caption{Baselines for a fixed amount of noise on top of user input.}\vspace{-0.1cm}
\label{table:baselines_exp1}
\centering
\resizebox{\columnwidth}{!}{%
\begin{tabular}{lcccccccc}
&
\multicolumn{2}{c}{Door opening}    &
\multicolumn{2}{c}{Tool use }   &
\multicolumn{2}{c}{In-hand man.}   &
\multicolumn{2}{c}{Object rel.}   \\
\cmidrule(lr){2-3}
\cmidrule(lr){4-5}
\cmidrule(lr){6-7}
\cmidrule(lr){8-9}
Method &
\multicolumn{1}{c}{Train}     &
\multicolumn{1}{c}{Test} &
\multicolumn{1}{c}{Train}     &
\multicolumn{1}{c}{Test} &
\multicolumn{1}{c}{Train}     &
\multicolumn{1}{c}{Test} &
\multicolumn{1}{c}{Train}     &
\multicolumn{1}{c}{Test} \\
 \midrule
 IK	& 64.00 & 74.00 &  50.00 & 56.00  & 67.67 & 69.92 & 77.00 & 83.00 \\
 RL-no user	& 75.00 & 59.00  & 51.00 & 44.00  & 43.61 & 38.34 & 0.00 &  0.00 \\
 IL-no user	& 0.00 & 0.00 &   0.00 & 0.00     & 4.00& 6.77 & 0.00  & 0.00  \\
 Hybrid-no res.& 0.00 & 0.00 &   0.00 & 0.00 & 4.00 & 0.00   & 0.00  & 0.00  \\
 RL+user reward	& 69.92 & 62.40  &   6.01  & 9.02 &48.12 & 27.81 & 0.00  & 0.00  \\
Hybrid+user rew. &  0.00 & 0.00 & 56.39 & 33.08  & 9.02 & 7.51 &  0.00 &  0.00  \\
\midrule
Ours & \textbf{81.33} & \textbf{83.00} &\textbf{ 61.00} &\textbf{ 58.00} & \textbf{90.97} & \textbf{87.21} &  \textbf{49.62} & \textbf{16.54} \\
\bottomrule
\end{tabular}}
\vspace{-0.4cm}
\end{table}
\subsubsection{Overcoming structured hand pose estimation and mapping errors}
In this experiment, we aim to verify that our algorithm can also deal with the structured noise injected via the hand pose estimator and the mapping function. We generate the training data using our strategy described in Section \ref{subsection:data_gen}. After creating the dataset, we also need to design a mapping function from the hand pose to the virtual model. Leveraging our data sampling strategy, we create pairs of data $(x_t, a_t)$. Other settings remain the same.

\textbf{Supervised IK baseline:} We use these pairs to train a function $\kappa(x_t)$ in a supervised setting. Our IK network is a (64, 64) MLP trained with a regression loss. In Table \ref{table:baselines_exp2} we observe that this function alone is not enough although it can achieve moderate success on the `door opening' scenario when ground-truth (GT, not noisy) poses are used.

\begin{table}[t]\vspace{0.2cm}
\caption{Baseline on structured hand pose error on ground-truth (GT) and estimated hand poses (Est.)}
\label{table:baselines_exp2} \vspace{-0.2cm}
\centering
\resizebox{0.85\columnwidth}{!}{%
\begin{tabular}{lcccc}
&
\multicolumn{2}{c}{Door opening} &
\multicolumn{2}{c}{In-hand man.}  \\
\cmidrule(lr){2-3}
\cmidrule(lr){4-5}
Method (Training set) &
\multicolumn{1}{c}{GT}     &
\multicolumn{1}{c}{Est.} &
\multicolumn{1}{c}{GT.}     &
\multicolumn{1}{c}{Est.} \\
\midrule
IK & 49.62 & 27.81 &   0.00 & 20.30 \\
RL - no user (GT) & 98.49 & 76.69 &  13.53 & 25.56 \\
RL - no user (Est.)	& 66.16 &   71.42 & 13.53 & 0.00  \\
RL + user reward (GT) & 0.00 & 0.00 &  \textbf{45.86} & 32.33 \\
RL + user reward (Est.)	& 0.00 & 0.00 &  0.00 & 12.03 \\
\midrule
Ours (Experiment A.1)	& 57.14 & 38.34 &  10.52 & 0.00 \\
Ours (GT poses)	& 83.45 & 42.10 & 10.52 & 32.33 \\
Ours (Est. poses) & \textbf{85.95} & \textbf{70.67} &  20.33 & \textbf{57.14}\\
\bottomrule
\end{tabular}}
\vspace{-0.5cm}
\end{table}

\begin{figure}[!htp]
\begin{center}
  \includegraphics[width=0.90\columnwidth, trim = {11.5cm 12cm 0cm 0cm},clip]{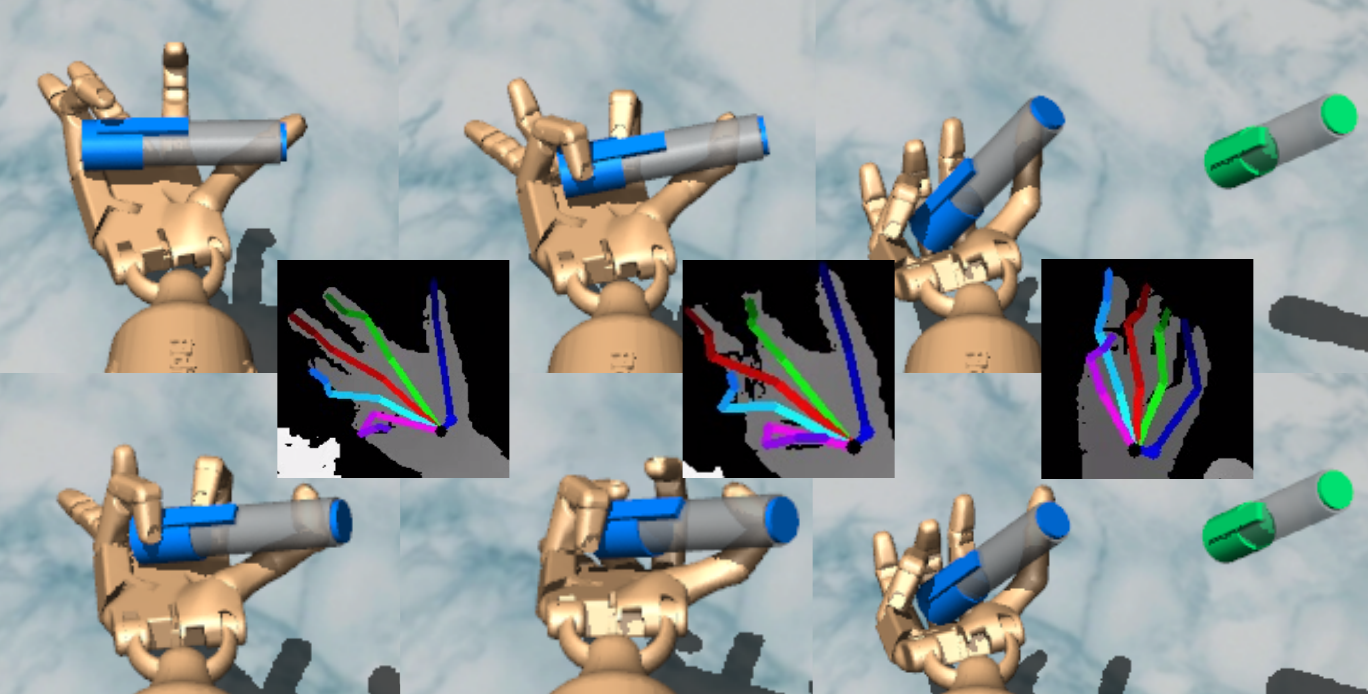}
\end{center}
\vspace{-0.4cm}
\caption{Qualitative results on `in-hand manipulation task'. (Middle) estimated hand pose (Top) IK result (Bottom) Our result. Depth images are retrieved using our data generation scheme.}
\label{fig:exp2}
\vspace{-0.3cm}
\end{figure}

In Table \ref{table:baselines_exp2} the results of both the best performing baselines in the previous experiment and our algorithm are depicted. We show results for both ground-truth poses and estimated (Est.) hand poses passed through IK.  We observe that our approach can achieve the task even when the IK output is poor (`in-hand') and offers solid performance when we observe better inputs (door). Using RL with user-augmented reward improves the IK baseline on `in-hand', however it struggles when noise from hand pose estimator is added. `RL-no user' performs well on `door-op', however in this baseline the virtual hand does not follow the user input and acts independently, similarly to triggering a prerecorded sequence. In Fig. \ref{fig:exp2} we show qualitative results on `in-hand' and in Fig. \ref{fig:exp1} (c) generated contact forces. Applying our models trained on the previous experiment did not perform well due to the different noise nature between both experiments, motivating our data generation scheme. 
\vspace{-0.1cm}

\subsection{Physics-based hand-object sequence reconstruction} \label{exp:B}
\begin{figure*}[!htp]
\vspace*{0.2cm}
\begin{center}
  \includegraphics[width=1\textwidth, trim = {0cm 7.8cm 5.3cm 0cm},clip]{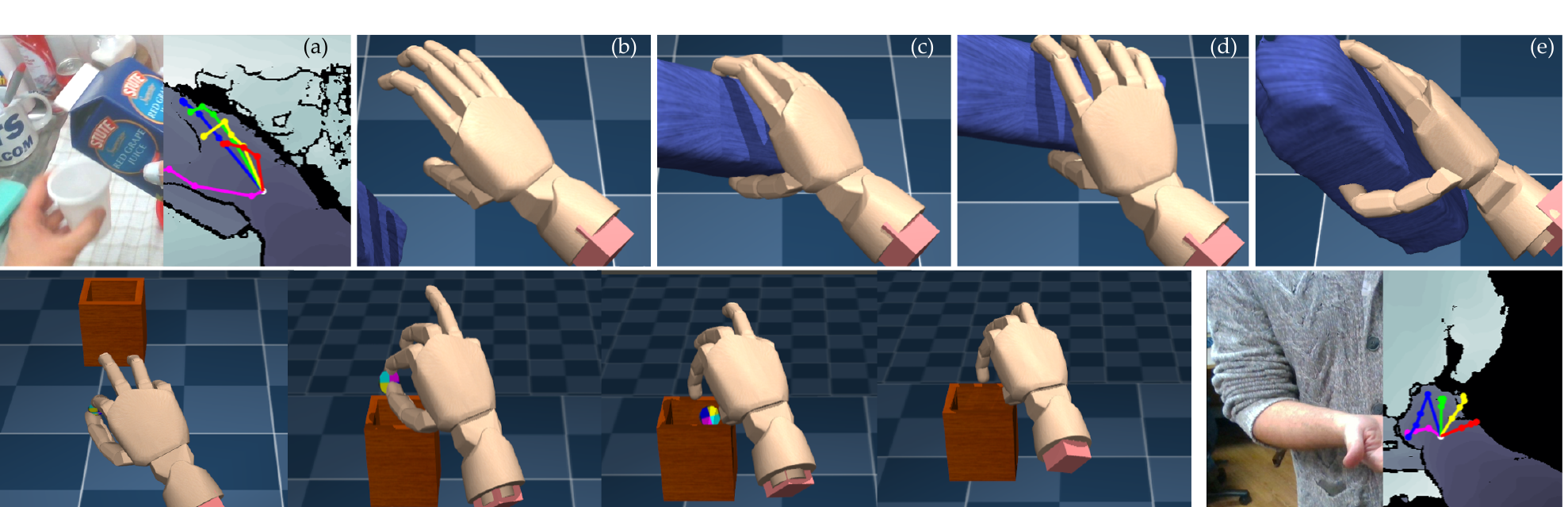}
\end{center}
\vspace{-0.3cm}

\caption{Qualitative results. Top row: a frame belonging to `pouring juice' action from F-PHAB dataset and its reconstruction for different methods from a fixed camera viewpoint. (a) RGB/depth image and estimated 3D hand pose. (b) IK function $\kappa$ \cite{antotsiou2018task} on HPE. (c) Closing hand baseline on top of $\kappa$. (d) Our approach without visual features and pose reward. (e) Our full approach, it produces a hand posture closer to the one depicted by the reference visual hand motion. Bottom row: Qualitative result on a `give coin' sequence. The task is achieved when the coin is placed on the other person's hand (red box).}
\label{fig:exp3}
\vspace{-0.6cm}
\end{figure*}
\label{subsec:exp3}
In this experiment we test our framework on the challenging task of transferring hand-object interactions from the real visual domain to a physically accurate simulation space. As a testbed, we use the First-Person Hand Action Benchmark (F-PHAB) \cite{garcia2018first}, providing hand-object interaction sequences with hand and object pose annotations. We select two different manipulation tasks covering two extreme cases of power and precision grasps: `pour juice from a carton to a glass' and `give coin to other person'. Each task contains 24 annotated video sequences from 6 different users and we use the 1:1 split of \cite{garcia2018first} for train-test data partition.

We recreate the real environment on the virtual space by placing a virtual object that we initialise with the 6D ground-truth pose. For the coin environment, we also place a target box that simulates `the hand of the other person'. We build the environments on MuJoCo and use the MPL hand model that consists of 23 DoF + 6 DoF \cite{kumar2015mujoco} and $\omega^{\mathsf{pose}}$ value of 0.01. As 3D hand pose estimator we use DeepPrior++ \cite{oberweger2017deepprior++}, extracting visual features $\phi_t \in \mathbb{R}^{1024}$ from the FC2 layer; and trained on the full dataset following the same 1:1 protocol which yields an average test joint error of 14.54 mm. Note that in this setup we do not have access to expert demonstrations, thus we cannot compute $r^{\mathsf{il}}$ nor use our data generation scheme. The rest of network architectures and parameters are the same as in previous experiment.

\textbf{Baselines:} In this experiment we implement two baselines, an IK function $\kappa$ following \cite{antotsiou2018task} and a `closing' baseline that acts on top of $\kappa$ and attempts to tighten the grasp or generate more contact forces similar to \cite{holl2018efficient}. 

\textbf{Metrics:} We use three different criteria to measure performance. First, `task success' measures the percentage of the times that the interaction is successful on test sequences. 
$\mathbf{E}_{\mathsf{pose}}$ measures the 3D hand pose error, in mm, by reprojecting $z_t$ to the input RGBD image space and comparing to ground-truth annotations, which gives us a notion on how similar the virtual posture looks compared to the actual visual pose. $\bar{T}$ measures the average length (in percentage over the total length) of the sequence before the simulation becomes unstable and the task is not completed successfully.

In Table~\ref{table:baselines_exp3} we show quantitative results on `pour juice' and `give coin' actions. We observe that our approach is able to accomplish the task while keeping a hand posture similar to the visual input (qualitative results are shown in Fig. \ref{fig:exp3}) and perform better than all baselines at train and test time. We observe that introducing the pose reward encourages a virtual pose closer to the visual input. Note that the virtual model is fixed in terms of bone lengths and kinematics and thus the reprojected pose will have an error offset even if the mapping was perfect. We show a successful example with generated contact forces in Fig.~\ref{fig:exp1} (c)). We observe a significant gap between training and test results that is even more severe in the `give coin' scenario where all the baselines show poor results in both training and test sets. Slight inaccuracies make the light and thin coin fall and thus failing in the task. We suspect that there are two main reasons for this. First, hand pose estimation errors are more severe than in the previous experiment and propagate through the hand model. Second the small number of training sequences may lead our network to overfit to the training set to some extent. This effect could be relieved by recording more training data or some data/trajectory augmentation technique. Note that the results on training sequences are still meaningful in the problem of offline motion reconstruction \cite{peng2018sfv}. 
\begin{table}[t]
\caption{Hand-object reconstruction of sequences in-the-wild}
\label{table:baselines_exp3}
\centering
\resizebox{\columnwidth}{!}{%
\begin{tabular}{lcccccc}
& \multicolumn{3}{c}{Training} & \multicolumn{3}{c}{Test} \\
\cmidrule(lr){2-4} \cmidrule(lr){5-7}
Method (Pour Juice) &
\multicolumn{1}{c}{$\bar{T}\uparrow$} &
\multicolumn{1}{c}{$\mathbf{E}_{\mathsf{pose}}\downarrow$} &
\multicolumn{1}{c}{Success $\uparrow$}     &
\multicolumn{1}{c}{$\bar{T}\uparrow$}     &
\multicolumn{1}{c}{$\mathbf{E}_{\mathsf{pose}}\downarrow$} &
\multicolumn{1}{c}{Success $\uparrow$}\\ 
 \midrule
 IK \cite{antotsiou2018task} & 18.0 & 26.95 &  16.0  & 24.8 &  33.22 & 5.0  \\
 Closing hand & 85.4 &  24.78 &55.0 & 47.0 & 35.46 & 38.0  \\
 Ours w/o pose reward & 97.4 & 26.82  & 84.0 & 52.0 &  37.88 & 47.0 \\
 \midrule
 Ours & \textbf{98.2} & 25.43 & \textbf{93.0} & \textbf{59.6} & \textbf{33.15}& \textbf{65.0}  \\
\toprule
Method (Give coin) &
\multicolumn{1}{c}{$\bar{T}\uparrow$}     &
\multicolumn{1}{c}{$\mathbf{E}_{\mathsf{pose}}\downarrow$} &
\multicolumn{1}{c}{Success $\uparrow$}     &
\multicolumn{1}{c}{$\bar{T}\uparrow$}     &
\multicolumn{1}{c}{$\mathbf{E}_{\mathsf{pose}}\downarrow$} &
\multicolumn{1}{c}{Success $\uparrow$}     \\
\midrule
 IK \cite{antotsiou2018task} & 9.2 & 24.90 &  0.0  &11.5 &  25.93 & 0.0  \\
 Closing hand & 55.4 &  28.44 & 25.0 & 70.2  & 33.70 & 28.57  \\
 \midrule
 Ours & \textbf{95.5} & \textbf{24.3} & \textbf{80.0} & \textbf{92.1}& \textbf{25.30} & \textbf{83.3} \\
 \bottomrule
 \end{tabular}
}
\vspace{-0.5cm}
\end{table}
\vspace{-0.1cm}
\section{CONCLUSION AND FUTURE WORK}
\vspace{-0.1cm}
We presented a framework that can perform dexterous manipulation skills by simply using a hand pose estimator without the need of any costly hardware. A residual agent learns within a physics simulator how to improve the user input to achieve a task while keeping the motion close to the input and expert recorded trajectories. We showed that our approach can be applied on two applications that require accurate hand-object motion while using noisy input poses.

We believe this paper can inspire future work and it can be extended in several different ways. For instance, making the full framework end-to-end, where the gradients propagate from the simulator to the hand pose estimator, is a promising direction for physics-based pose estimation. For the second application, it would also be interesting to add a 6D object pose estimator in the loop~\cite{sahin2020review}. Besides, generating synthetic data to close the training loop has also potential, for instance by fitting a realistic hand model in a similar way as in \cite{Armagan2020} on top of mocap data or already trained policies \cite{rajeswaran2017learning}. This could also help to narrow, to some extent, the training-test gap found in our experiments and make possible the deployment of the system to receive poses in a stream in a VR system and may prompt additional challenges. The study of RL generalization to both in-the-wild scenarios and other tasks is an open research problem. New results in these areas would benefit the present work, because how to scale up  the number of tasks in the current framework is not clear. 
\vspace{-0.1cm}
\bibliographystyle{IEEEtran}
\bibliography{IEEEabrv,biblio}
\setcounter{section}{0}
\renewcommand{\thesection}{\arabic{section}} 
\begin{center}
\Large\textbf{Appendix}\\
\end{center}\vspace{0.2cm}

\section{Learning algorithm box}
The pseudocode for the training process of our framework can be found in Algorithm~\ref{alg:training}.
\begin{algorithm}[h]
\caption{Framework training with data generation scheme, PPO and GAIL}
\label{alg:training}
\begin{algorithmic}
\STATE{$\theta, v, \psi \leftarrow$ initialize policy, value function and discriminator weights}
\STATE{reset $\leftarrow$ true}
\WHILE{not done}
    \STATE{Generate policy rollouts:}
	\FOR{step$= 1,...,m$}
	    \IF{reset}
	        \STATE{$\tau \leftarrow$ Randomly select one expert sequence of length $T$}
	        \STATE{$\{x_t, \phi_t\} \leftarrow $Generate hand pose and visual features sequences with Algorithm \ref{alg:dataset_gen} and $\tau$}
	        \STATE{Generate $\kappa(x_t(\phi_t))$ and augment translation actions from $\tau$ with noise}
            \STATE{Initialize environment to state $s_0$ from $\tau$}
        \ENDIF
    	\STATE{$f_t \sim  \pi_\theta(f_t | s_t, \kappa(x_t(\phi_t)), \phi_t)$}
    	
        \STATE{$a_t \leftarrow \kappa(x_t(\phi_t))-f_t$}
    	\STATE Apply $a_t$ and simulate forward one step
    	\STATE{$s_{t+1} \leftarrow$ end state}
    	\STATE{$r_t \leftarrow \omega^{\mathsf{task}}r^{\mathsf{task}}_t+ \omega^{\mathsf{il}}r^{\mathsf{il}}_t+\omega^{\mathsf{pose}}r^{\mathsf{pose}}_t$}
    	\STATE{Record $(s_t, a_t, r_t, s_{t+1})$ into memory}
	\ENDFOR
    \STATE{Update $\theta$, $v$ and $\psi$ with their respective gradients and recorded batches by following \cite{schulman2017proximal} and \cite{ho2016generative}}
\ENDWHILE
\end{algorithmic}
\end{algorithm}
\section{Learning algorithm parameters}
We use the same learning parameters for both our method and the baselines. Our approach and Hybrid/IL baselines have three networks: policy network, value function network and discriminator. RL baselines have only policy and value networks. Policy updates are performed after a batch of $m=4096$ samples has been collected and minibatches of size $256$ are sampled for each gradient step. We use Adam optimizer with learning rate for the policy and value function of $3\cdot10^{-4}$ and $10^{-4}$ for the discriminator network. After a batch of data is collected 15 parameter updates are performed for both policy and value networks and 5 for the discriminator. PPO parameters \cite{schulman2017proximal} are as follows: $0.995$ for the temporal discount, $0.97$ for the GAE parameter and $0.2$ for the clipping threshold. The initial policy standard deviation is set to $0.01$. $\lambda$ is set to $0.5$ in all the experiments involving hybrid reward. Networks updates are performed on a single GPU (NVIDIA GTX 1080Ti) using TensorFlow and data samples are generated solely on CPU. For the supervised mapping function $\kappa(x_t)$ used in Experiment Section 4.1.2, we use a 2 layer MLP (64-64), Adam update $10^{-3}$, $32$ batch size and $10^{6}$ iterations. We optimize this mapping on the same GPU as above and using TensorFlow.

\section{Details on data generation scheme}
If we are given an expert action $a_t \sim \pi_E$ and we observe Eq. \ref{eq: action_res}, it holds that $f_{\theta}(s_t, \kappa(x_t(\phi_t)), \phi_t) \approx 0$. Given that $a_t$ has been recorded using a data glove, we can assume that the depicted pose from the virtual pose $z_t$ has a similar posture to $x_t$. We set a camera viewpoint in the simulation to be placed in a similar place as the depth sensor in the real space and compute the relative viewpoint $v_t$, i.e. elevation and azimuth angles, between this camera viewpoint and a normal vector from the $z_t$'s palm.  Note that both poses belong to different domains and thus are not directly comparable. To deal with this, we normalize $z_t$ to have all the joint links to be unit vectors, and rotate the palm to be aligned to a certain plane obtaining $\hat{z}_t$. We follow the same process on a hand pose dataset containing pairs of depth images and ground-truth hand pose annotations. We then query the dataset to first retrieve all the ground-truth hand poses that have the same viewpoint $v_t$, and retrieve the nearest neighbor by computing the Euclidean distance between $\hat{z}_t$ and the normalized set of ground-truth candidates.
 We then retrieve the associated depth image and compute the estimated hand pose.  We do not consider the translation of the hand in the image given that this seriously limits the number of candidate poses, however we deal with this by generating different positions by adding noise on the ground-truth translations, which also allows us to generate diverse realistic sequences, thus making our training more diverse. 
See Algorithm \ref{alg:dataset_gen}.

\begin{algorithm}[h]
\caption{Data generation scheme}
\label{alg:dataset_gen}
\begin{algorithmic}[1]
\REQUIRE{$\tau=\{s_t,a_t\} \in \mathcal{D}$ sequence of expert demonstrations of length $T$}
    \STATE{$s_0 \leftarrow$ sample initial state from $\tau$}
    \WHILE{$t<T$}
    \STATE{Apply $a_t$ to the environment}
    \STATE{$z_t \leftarrow$ read simulation sensors}
    \STATE{$v_t \leftarrow$ compute relative viewpoint between $z_t$'s 
    palm and simulator camera}
    \STATE{$\hat{z_t} \leftarrow$ normalize and align $z_t$}
    \STATE{$\hat{x}_t \leftarrow$ query dataset with $v_t$ and $\hat{z_t}$}
    \STATE{$x_t, \phi_t \leftarrow$ apply hand pose estimator to $\hat{x}_t$'s associated image}
    \ENDWHILE
\ENSURE{$\{x_t\}$, $\{\phi_t\}$ sequences of estimated hand poses and visual features}
\end{algorithmic}
\end{algorithm}

\section{Experiment A: Experiment Details}
\subsection*{Experiment A.1: details and additional results}
MoCap user demonstrations are sampled from the dataset provided in \cite{rajeswaran2017learning}. For each task we are given $24$ MoCap demonstrations and we split them in equal training and test sets. We use the original environments from  Rajeswaran \textit{et al.} \cite{rajeswaran2017learning} with the modification of adding user input and making the action residual. We keep the simulator, physics parameters as in the original environments\footnote{\url{https://github.com/aravindr93/hand_dapg}} and states and task rewards are described in Section \ref{sec:tasks_states}. 

\begin{figure*}[!htp]
\begin{center}
  \includegraphics[width=1\textwidth, trim = {0cm 13cm 1.8cm 0cm},clip]{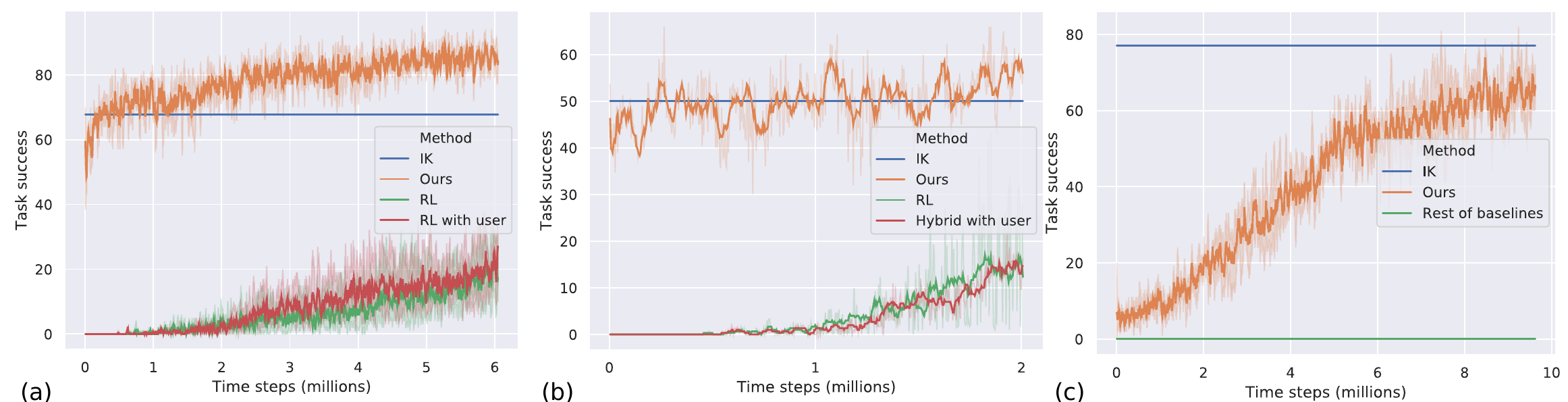}
\end{center}
\caption{Experiment A.1: Additional learning curves learning curves for our approach and the baselines that performed the best on \textbf{Table II}.  \textbf{(a)} In-hand manipulation task \textbf{ (b)} Tool use (hammer)  and \textbf{(c)} Object relocation. Plots are generated for three different random seeds and confidence interval  ($95\%$)}
\label{fig:exp1_additional}
\end{figure*}

Training and test data is generated by adding random Gaussian noise to these demonstrations. Reported train/test results are after averaging results for $100$ policy rollouts (augmented with noise) and three random seeds on learning algorithm. We provide additional learning curves for the rest of the tasks curves on Figure \ref{fig:exp1_additional}. We also show the learning curves for different levels of training noise (used to generate Table 1 of the main paper) in Figure \ref{fig:sigmas}.

\subsection*{Experiments A.2: Details and additional results}

We use the same training/test split as in previous experiment. BigHand2.2M dataset is queried for one user ($~2\cdot10^{5}$ samples) that has not been seen by the hand pose estimator in training. As a hand pose estimator we used the approach of \cite{ye2016spatial} trained on the rest of the subjects of BigHand2.2M. We report the following average hand pose estimation errors (Euclidean distance between annotation and estimation) for the different tasks, which are consistent with state-of-the-art results for depth images: 9.90 mm (door op.), 7.74 mm (in-hand man.), 8.76 mm (tool) and 7.42 mm (relocation).

For the dataset generation scheme the order of the query, i.e. first viewpoint or posture, will have an impact in the result. We empirically found that the best results were obtained with the following order: first query candidates based on azimuth, then proceed with altitude and end with pose distance.

\begin{figure*}[!htp]
\begin{center}
  \includegraphics[width=0.4\textwidth, trim = {0cm 11.0cm 15.0cm 0cm},clip]{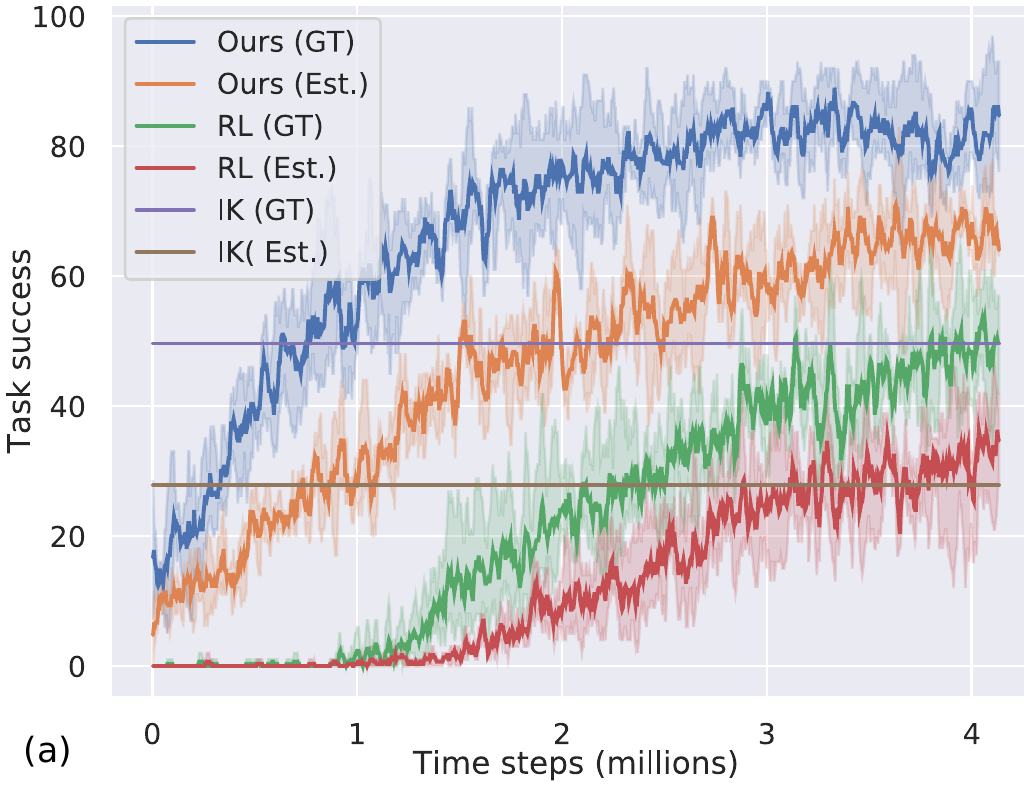}
\end{center}
\caption{Experiment A.2: training curves (door).}
\label{fig:exp2curves}
\end{figure*}

In Figure \ref{fig:exp2curves} we show learning curves for our approach and other baselines, we observe that our training converges faster and to a higher task success than the baselines. In Table \ref{table:baselines_exp2_supp} we expand the Table III from the main paper that we had to reduce for space reasons. Train and test split is generated as in previous experiment and the $x_t$ is generate following our data generation scheme. Noise of intensity $0.05$ rad is added only to the translation and rotation actuators of the arm from the   demonstrations, the rest of the user action (24) comes from $\kappa(x_t)$. Results are shown on generated test sequences for both ground-truth hand poses (GT) and estimated hand poses (Est.) on 100 policy rollouts for three different learning random seeds as in previous experiments. We observe that our approach outperforms all the baselines, except for the relocation environment, which is consistent with our result in A.1 and discussed in the main paper.

\section{Experiment A: states, rewards and tasks}
\label{sec:tasks_states}
The state space and the task rewards are the same as in the work of \cite{rajeswaran2017learning} with the addition of $\kappa(x)$ to the state and an user following reward (only on + user reward baselines). The action space is the same as in \cite{rajeswaran2017learning} for each task.
\subsection*{Door opening}
The user has to undo the latch before the door can be opened. The latch has a significant dry friction and a bias torque that forces the door to be closed. The success measure is defined as $door_{joint} > 1.0$ at the end of the interaction. The state is defined as
\begin{equation*}
s_{\mathsf{door}} = [hand_{joints}; palm_{pos}; door_{handle~pos, latch, hinge}]
\end{equation*}
and the reward as:
\begin{equation*}
\begin{split}
  r_{\mathsf{door}} = &10\mathbf{I}(door_{pos}>1.35) + 8\mathbf{I}(door_{pos}>1.0)\\
            & + 2\mathbf{I}(door_{pos}>1.2) - 0.1||door_{pos}-1.57||_{2}.\\
\end{split}
\end{equation*}

\subsection*{Tool Use: Hammer}
We consider using a hammer to drive in a nail. The user hand has to pick up the hammer from the ground, move it over to the nail and hammer in with a significant force to get the nail to move into the board. The nail has dry friction capable of absorbing up of 15N of force. There are more than one steps needed to perform this task, which require accurate grasping and positioning.  The success measure is defined as $||nail_{pos}-nail_{pos}^{goal}||_{2}<0.01$ at the end of the interaction. The state is defined as:

\begin{equation*}\begin{split}
s_{\mathsf{hammer}} = [hand_{joints, velocity}; palm_{pos}; hammer_{pos, rot};\\
nail_{pos}^{goal}; nail_{impact force}]
\end{split}\end{equation*}
and the reward as:
\begin{equation*}
\begin{split}
  r_{\mathsf{hammer}} = & 75 * \mathbf{I}(||nail_{pos}^{goal}-nail_{pos}||_{2}<0.10) + \\
  	  & 25 * \mathbf{I}(||nail_{pos}^{goal}-nail_{pos}||_{2}<0.02) - \\
      & 10||nail_{pos}^{goal}-nail_{pos}||_{2}.\\
\end{split}
\end{equation*}

\subsection*{In-hand Manipulation: Repositioning a pen} 
The user goal is to reposition the blue pen to a desired target orientation in-hand, visualized by the green pen. The base of the hand is fixed. The pen is highly underactuated and requires careful application of forces by the hand to reposition it. Most actions lead to catastrophic failure like dropping the object. The success measure is  $||pen_{rot}-pen_{rot}^{goal}||_{cosine}>0.95)$ at the end of the interaction. The state is defined as
\begin{equation*}
s_{\mathsf{pen}} = [hand_{joints}; pen_{pos, rot}; pen_{pos, rot}^{goal}]
\end{equation*}
and the reward as:
\begin{equation*}
\begin{split}
  r_{\mathsf{pen}} = &50( \mathbf{I}(||pen_{pos}^{goal}-pen_{pos}||_{2}<0.075) \otimes \\ 
              & \mathbf{I}(||pen_{rot}-pen_{rot}^{goal}||_{cosine}>0.95)).\\  
\end{split}
\end{equation*}
\subsection*{Object relocation}
The user goal is to use the hand to pick up the blue ball and move it to the green target location. The success measure is $||object_{pos}-object_{pos}^{goal}||_{2}<0.05)$ at the end of the interaction. The state is defined as:

\begin{equation*}
s_{\mathsf{relo}} = [hand_{joints}; palm_{pos}, object_{pos}; object_{pos}^{goal}]
\end{equation*}
and the reward is defined as:
\begin{equation*}
\begin{split}
  r_{\mathsf{relo}} = &10\mathbf{I}(||object_{pos}-object_{pos}^{goal}||_{2}<0.1) + \\ 
       & 20\mathbf{I}(||object_{pos}-object_{pos}^{goal}||_{2}<0.05).\\ 
\end{split}
\end{equation*}

\subsection*{User reward (+ user reward baselines only)}
We design a reward function to encourage the agent to follow user action and add it to the reward functions presented above for the RL+user reward and Hybrid+user reward baselines. It is defined as:
\begin{equation*}
  r_{user} = -0.1||a-\kappa(x)||_{2}.
\end{equation*}

\section{Experiment B: Experiment Details}
Networks and hyperparameters are the same as described above. As hand pose estimator we use DeepPrior++~\cite{oberweger2017deepprior++} trained with the protocol described in \cite{garcia2018first}. Visual features $\phi$ are extracted from the last full connected layer with dimension $1024$. The environments are implemented in MuJoCo using an extended version of the MPL model~\cite{kumar2015mujoco} by \cite{antotsiou2018task}, we query the policy at 30 Hz and the simulator at 200 Hz. We use the inverse kinematics function from \cite{antotsiou2018task} to do the first mapping between estimated hand poses and hand model. We control the 23 actuators of the hand model, while the global rotation and orientation of the model in the virtual space are predicted by the hand pose estimator. As action representation, PD controllers are used to compute joint torques with default gain parameters~\cite{kumar2015mujoco}. In this experiment we found that using early termination (i.e. resetting the environment when the object falls far away) helped to make the training converge faster. 25\% of the action space is used as residual domain.

\subsection*{Pouring juice action}
This task consists of holding a juice bottle and making the pour action by following the user input. The environment has only one object consisting of the juice bottle. The task is considered successful if at the end of the clip the virtual model is holding the bottle. There are 24 sequences of about 100 frames each and we use 12 for training and 12 for test.
The state is defined as:
\begin{equation*}
\begin{split}
s_{\mathsf{juice}} = [hand_{joints}, hand_{vel}, object_{pos}-hand_{pos},\\
object_{rot}-hand_{rot}, contact],
\end{split}
\end{equation*}

where $contact$ is a function that measures the normalized distance between contact points between finger tips and object surface similar to the PSO fit function of \cite{antotsiou2018task}. We place a sensor to measure contacts at each finger tip, and thus $contacts$ has a value for each fingertip (5-D). The input to the policy network is $1104$ (this number includes the state, the visual features and the user input).

\begin{equation*}
\begin{split}
  r_{\mathsf{juice}} = \mathbf{I}(\sum contacts > 0)(1-\sum contacts)-\\||object_{pos}-hand_{pos}||_{2}
\end{split}
\end{equation*}

\textbf{Closing hand baseline}: This baseline is implemented by enforcing contact between object surface and finger tips by first reading user input and moving the actuators towards the object. We use this baseline as user input to our residual policy.

\subsection*{Give coin action}
The task consist of the user placing a coin on other user's hand. The virtual environment has three components: hand model, coin object and box. The box represents the other user's hand and the task is to place the coin within the limits of the box. There are two big challenges in this task: holding the small coin and carefully placing it in the box. In the original dataset there are 25 sequences, however we had to discard five of them due to simulator instabilities.

The state is defined as:
\begin{equation*}
\begin{split}
s_{\mathsf{coin}} = [hand_{joints}, hand_{vel}, coin_{pos}-index^{tip}_{pos},\\ coin_{pos}-thumb^{tip}_{pos}, box_{pos},box_{pos}-index^{tip}_{pos},\\
box_{pos}-thumb^{tip}_{pos}, box_{pos}-coin_{pos},\\ coin_{rot}-hand_{rot}, contact],
\end{split}
\end{equation*}
where in this case $contacts$ is limited to index and thumb finger tips. The input to the policy network is $1119$.

The reward for this task consists of two parts that depend on the two main stages of the action: holding the coin and carefully placing it in the box. The reward function for this task is defined as follows:

\begin{equation*}
\begin{split}
  r_{\mathsf{coin}}=\mathbf{I}(d(box_{pos},index^{tip}_{pos})>0.12\land  d(box_{pos},coin_{pos})>0.05)\times \\
  (1-\sum contacts)-d(coin_{pos},thumb^{tip}_{pos})-d(coin_{pos},index^{tip}_{pos}))+\\
  \mathbf{I}(d(box_{pos},index^{tip}_{pos})\leq 0.12 \lor d(box_{pos},coin_{pos})\leq0.05)\times\\
  (1.5\mathbf{I}(d(box_{pos},coin_{pos})<0.05)+\mathbf{I}(d(box_{pos},coin_{pos})<0.08)-\\
  d(box_{pos},index^{tip}_{pos})),
\end{split}
\end{equation*}

where $d(\cdot,\cdot)$ represents the Euclidean distance between two bodies. The first two lines of the equation represent the `approaching box' phase where contact between index, thumb and coin are encouraged. The last two lines give a high reward when the coin is within the limits of the box and penalizes when the coin is far from the target (e.g. the coin fell outside the box).

\textbf{Closing hand baseline:} In this task we move the index and thumb fingers towards the coin to make a `pinch' gesture. Once the hand is near the box, the fingers release the coin making also a subtle wrist movement. We use this baseline as user input to our residual policy.

\section{Qualitative results}
Videos can be visualised on the project webpage:

\url{https://sites.google.com/view/dexres}

\begin{figure*}[!htp]
\begin{center}
  \includegraphics[width=0.4\textwidth, trim = {0cm 23.5cm 13.5cm 0cm},clip]{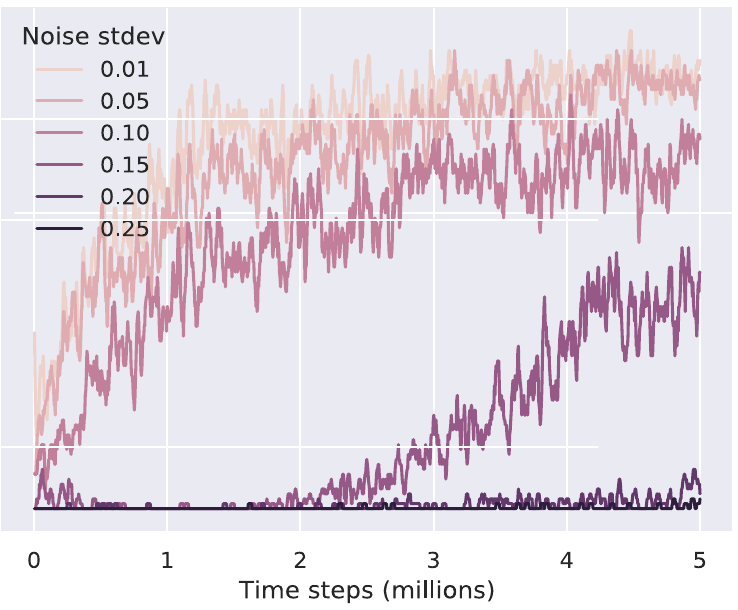}
\end{center}
\caption{Training curves for different levels of noise in training. These models were used to generate Table I in the main paper.}
\label{fig:sigmas}
\end{figure*}

\begin{table*}[ht]
   \caption{\textbf{Experiment A.2}: Baseline comparison for all baselines and tasks (expands Table III)}
     \label{table:baselines_exp2_supp}
   \centering
  \begin{tabular}{lcccccccc}
&
\multicolumn{2}{c}{Door opening}    &
\multicolumn{2}{c}{In-hand man.}   &
\multicolumn{2}{c}{Tool use (hammer)}   &
\multicolumn{2}{c}{Object relocation}  \\
\cmidrule(lr){2-3}
\cmidrule(lr){4-5}
\cmidrule(lr){6-7}
\cmidrule(lr){8-9}

Method (Training set)&
\multicolumn{1}{c}{GT}     &
\multicolumn{1}{c}{Est.} &
\multicolumn{1}{c}{GT}     &
\multicolumn{1}{c}{Est.} &
\multicolumn{1}{c}{GT}     &
\multicolumn{1}{c}{Est.} &
\multicolumn{1}{c}{GT.}     &
\multicolumn{1}{c}{Est.} \\
 \midrule
  IK 						 & 49.62 & 27.81 &   0.00 & 20.30 & 66.16 & 68.42 & 82.70 & 90.22\\
  RL - no user (GT)						 & 98.49 & 76.69 &  13.53 & 25.56 & 34.59 & 29.32 & 0.00 & 0.00 \\
  IL - no user (GT)						 & 0.00 & 0.00 &  0.00 & 0.00 & 0.00 & 0.00 & 0.00 & 0.00 \\
  Hybrid - no user (GT)						 & 0.00 & 0.00 &  20.30 & 9.02 & 39.84 & 37.59 & 0.00 & 0.00 \\
    RL - no user (Est.)						 & 66.16 &   71.42 & 13.53 & 0.00 & 58.65 & 54.89 & 0.00 & 0.00 \\
  IL - no user (Est.)						 & 0.00 &   0.00 & 0.00 & 0.00 & 0.00 & 0.00 & 0.00 & 0.00 \\

    Hybrid  - no user (Est.)					 &  0.00 &  0.00 & 12.03 & 10.52 & 53.38 & 47.37 & 0.00 & 0.00 \\
  \midrule
  RL + user reward (GT)						 & 0.00 & 0.00 &  \textbf{45.86} & 32.33 & 3.76 & 3.76 & 0.00 & 0.00\\
    Hybrid + user reward (GT)						 & 0.00 & 0.00 &  0.00 &  12.03 & 58.64 & 29.32 & 0.00 & 0.00\\
  RL + user reward (Est.)						 & 0.00 & 0.00 &  0.00 & 12.03 & 12.78 & 4.51 & 0.00 & 0.00 \\
    Hybrid + user reward (Est. )						 & 0.00 & 0.00 &  0.00 & 0.00 & 54.13 & 68.00 & 0.00 & 0.00\\
    \midrule
    Ours (Experiment A.1)						 & 57.14 & 38.34 &  10.52 & 0.00 & 60.15 & 30.82 & 21.80 & 29.32 \\
     Ours (GT poses)						 & 83.45 & 42.10 & 10.52 & 32.33 & 78.00 & 25.56 & 34.00 & 12.78\\
  Ours (Est. poses)						 & \textbf{85.95} & \textbf{70.67} &  20.33 & \textbf{57.14} & \textbf{78.94} & \textbf{71.42} & 34.00 & 35.00\\
  \bottomrule
  \end{tabular}
\end{table*}

\end{document}